\documentclass{Interspeech2024}
\interspeechcameraready

\usepackage{footnote}
\makesavenoteenv{tabular}
\makesavenoteenv{table}
\usepackage{caption}
\usepackage{subcaption}
\usepackage{todonotes}
\usepackage{hyperref}
\usepackage[capitalize]{cleveref}
\usepackage{amsmath}
\usepackage[acronym, shortcuts, nohypertypes={acronym}]{glossaries}
\newacronym{ASR}{ASR}{automatic speech recognition}
\newacronym{CNN}{CNN}{convolutional neural network}
\newacronym{CRNN}{CRNN}{convolutional recurrent neural network}
\newacronym{DL}{DL}{deep learning}
\newacronym{DNN}{DNN}{deep neural network}
\newacronym{GMM}{GMM}{Gaussian mixture model}
\newacronym{LSTM}{LSTM}{long short-term memory recurrent neural network}
\newacronym{MAC}{MAC}{multiply-addition count}
\newacronym{ML}{ML}{machine learning}
\newacronym{MLP}{MLP}{multi-layered perceptron}
\newacronym{NLP}{NLP}{natural language processing}
\newacronym{PANN}{PANN}{pretrained audio neural network}
\newacronym{RNN}{RNN}{recurrent neural network}
\newacronym{SER}{SER}{speech emotion recognition}
\newacronym{SSL}{SSL}{self-supervised learning}
\newacronym{SVM}{SVM}{support vector machine}
\newacronym{TDNN}{TDNN}{time delay neural network}
\newacronym{TL}{TL}{transfer learning}
\newacronym{UAR}{UAR}{unweighted average recall}

\usepackage{xcolor}

\usepackage{colortbl}
\PassOptionsToPackage{dvipsnames,table}{color}
\usepackage{pgfplotstable}
\pgfplotsset{compat=1.14}
\definecolor{b0}{rgb}{1.0000, 1.0000, 1.000}
 \definecolor{b1}{rgb}{0.990, 1.0000, 1.000}
 \definecolor{b2}{rgb}{0.980, 1.0000, 1.000}
 \definecolor{b3}{rgb}{0.970, 1.0000, 1.000}
 \definecolor{b4}{rgb}{0.960, 1.0000, 1.000}
 \definecolor{b5}{rgb}{0.950, 1.0000, 1.000}
 \definecolor{b6}{rgb}{0.940, 1.0000, 1.000}
 \definecolor{b7}{rgb}{0.930, 1.0000, 1.000}
 \definecolor{b8}{rgb}{0.920, 1.0000, 1.000}
 \definecolor{b9}{rgb}{0.910, 1.0000, 1.000}
\definecolor{b10}{rgb}{0.900, 1.0000, 1.000}
\definecolor{b11}{rgb}{0.890, 1.0000, 1.000}
\definecolor{b12}{rgb}{0.880, 1.0000, 1.000}
\definecolor{b13}{rgb}{0.870, 1.0000, 1.000}
\definecolor{b14}{rgb}{0.860, 1.0000, 1.000}
\definecolor{b15}{rgb}{0.850, 1.0000, 1.000}
\definecolor{b16}{rgb}{0.840, 1.0000, 1.000}
\definecolor{b17}{rgb}{0.830, 1.0000, 1.000}
\definecolor{b18}{rgb}{0.820, 1.0000, 1.000}
\definecolor{b19}{rgb}{0.810, 1.0000, 1.000}
\definecolor{b20}{rgb}{0.800, 1.0000, 1.000}
\definecolor{b21}{rgb}{0.790, 1.0000, 1.000}
\definecolor{b22}{rgb}{0.780, 1.0000, 1.000}
\definecolor{b23}{rgb}{0.770, 1.0000, 1.000}
\definecolor{b24}{rgb}{0.760, 1.0000, 1.000}
\definecolor{b25}{rgb}{0.750, 1.0000, 1.000}
\definecolor{b26}{rgb}{0.740, 1.0000, 1.000}
\definecolor{b27}{rgb}{0.730, 1.0000, 1.000}
\definecolor{b28}{rgb}{0.720, 1.0000, 1.000}
\definecolor{b29}{rgb}{0.710, 1.0000, 1.000}
\definecolor{b30}{rgb}{0.700, 1.0000, 1.000}
\definecolor{b31}{rgb}{0.690, 1.0000, 1.000}
\definecolor{b32}{rgb}{0.680, 1.0000, 1.000}
\definecolor{b33}{rgb}{0.670, 1.0000, 1.000}
\definecolor{b34}{rgb}{0.660, 1.0000, 1.000}
\definecolor{b35}{rgb}{0.650, 1.0000, 1.000}
\definecolor{b36}{rgb}{0.640, 1.0000, 1.000}
\definecolor{b37}{rgb}{0.630, 1.0000, 1.000}
\definecolor{b38}{rgb}{0.620, 1.0000, 1.000}
\definecolor{b39}{rgb}{0.610, 1.0000, 1.000}
\definecolor{b40}{rgb}{0.600, 1.0000, 1.000}
\definecolor{b41}{rgb}{0.590, 1.0000, 1.000}
\definecolor{b42}{rgb}{0.580, 1.0000, 1.000}
\definecolor{b43}{rgb}{0.570, 1.0000, 1.000}
\definecolor{b44}{rgb}{0.560, 1.0000, 1.000}
\definecolor{b45}{rgb}{0.550, 1.0000, 1.000}
\definecolor{b46}{rgb}{0.540, 1.0000, 1.000}
\definecolor{b47}{rgb}{0.530, 1.0000, 1.000}
\definecolor{b48}{rgb}{0.520, 1.0000, 1.000}
\definecolor{b49}{rgb}{0.510, 1.0000, 1.000}
\definecolor{b50}{rgb}{0.500, 1.0000, 1.000}
\definecolor{b51}{rgb}{0.490, 1.0000, 1.000}
\definecolor{b52}{rgb}{0.480, 1.0000, 1.000}
\definecolor{b53}{rgb}{0.470, 1.0000, 1.000}
\definecolor{b54}{rgb}{0.460, 1.0000, 1.000}
\definecolor{b55}{rgb}{0.450, 1.0000, 1.000}
\definecolor{b56}{rgb}{0.440, 1.0000, 1.000}
\definecolor{b57}{rgb}{0.430, 1.0000, 1.000}
\definecolor{b58}{rgb}{0.420, 1.0000, 1.000}
\definecolor{b59}{rgb}{0.410, 1.0000, 1.000}
\definecolor{b60}{rgb}{0.400, 1.0000, 1.000}
\definecolor{b61}{rgb}{0.390, 1.0000, 1.000}
\definecolor{b62}{rgb}{0.380, 1.0000, 1.000}
\definecolor{b63}{rgb}{0.370, 1.0000, 1.000}
\definecolor{b64}{rgb}{0.360, 1.0000, 1.000}
\definecolor{b65}{rgb}{0.350, 1.0000, 1.000}
\definecolor{b66}{rgb}{0.340, 1.0000, 1.000}
\definecolor{b67}{rgb}{0.330, 1.0000, 1.000}
\definecolor{b68}{rgb}{0.320, 1.0000, 1.000}
\definecolor{b69}{rgb}{0.310, 1.0000, 1.000}
\definecolor{b70}{rgb}{0.300, 1.0000, 1.000}
\definecolor{b71}{rgb}{0.290, 1.0000, 1.000}
\definecolor{b72}{rgb}{0.280, 1.0000, 1.000}
\definecolor{b73}{rgb}{0.270, 1.0000, 1.000}
\definecolor{b74}{rgb}{0.260, 1.0000, 1.000}
\definecolor{b75}{rgb}{0.250, 1.0000, 1.000}
\definecolor{b76}{rgb}{0.240, 1.0000, 1.000}
\definecolor{b77}{rgb}{0.230, 1.0000, 1.000}
\definecolor{b78}{rgb}{0.220, 1.0000, 1.000}
\definecolor{b79}{rgb}{0.210, 1.0000, 1.000}
\definecolor{b80}{rgb}{0.200, 1.0000, 1.000}
\definecolor{b81}{rgb}{0.190, 1.0000, 1.000}
\definecolor{b82}{rgb}{0.180, 1.0000, 1.000}
\definecolor{b83}{rgb}{0.170, 1.0000, 1.000}
\definecolor{b84}{rgb}{0.160, 1.0000, 1.000}
\definecolor{b85}{rgb}{0.150, 1.0000, 1.000}
\definecolor{b86}{rgb}{0.140, 1.0000, 1.000}
\definecolor{b87}{rgb}{0.130, 1.0000, 1.000}
\definecolor{b88}{rgb}{0.120, 1.0000, 1.000}
\definecolor{b89}{rgb}{0.110, 1.0000, 1.000}
\definecolor{b90}{rgb}{0.100, 1.0000, 1.000}
\definecolor{b91}{rgb}{0.090, 1.0000, 1.000}
\definecolor{b92}{rgb}{0.080, 1.0000, 1.000}
\definecolor{b93}{rgb}{0.070, 1.0000, 1.000}
\definecolor{b94}{rgb}{0.060, 1.0000, 1.000}
\definecolor{b95}{rgb}{0.050, 1.0000, 1.000}
\definecolor{b96}{rgb}{0.040, 1.0000, 1.000}
\definecolor{b97}{rgb}{0.030, 1.0000, 1.000}
\definecolor{b98}{rgb}{0.020, 1.0000, 1.000}
\definecolor{b99}{rgb}{0.010, 1.0000, 1.000}
\definecolor{b100}{rgb}{0.000, 1.0000, 1.000}

\definecolor{p0}{rgb}{1.000, 1.000, 1.000}
\definecolor{p1}{rgb}{1.000, 0.990, 1.000}
\definecolor{p2}{rgb}{1.000, 0.980, 1.000}
\definecolor{p3}{rgb}{1.000, 0.970, 1.000}
\definecolor{p4}{rgb}{1.000, 0.960, 1.000}
\definecolor{p5}{rgb}{1.000, 0.950, 1.000}
\definecolor{p6}{rgb}{1.000, 0.940, 1.000}
\definecolor{p7}{rgb}{1.000, 0.930, 1.000}
\definecolor{p8}{rgb}{1.000, 0.920, 1.000}
\definecolor{p9}{rgb}{1.000, 0.910, 1.000}
\definecolor{p10}{rgb}{1.000, 0.900, 1.000}
\definecolor{p11}{rgb}{1.000, 0.890, 1.000}
\definecolor{p12}{rgb}{1.000, 0.880, 1.000}
\definecolor{p13}{rgb}{1.000, 0.870, 1.000}
\definecolor{p14}{rgb}{1.000, 0.860, 1.000}
\definecolor{p15}{rgb}{1.000, 0.850, 1.000}
\definecolor{p16}{rgb}{1.000, 0.840, 1.000}
\definecolor{p17}{rgb}{1.000, 0.830, 1.000}
\definecolor{p18}{rgb}{1.000, 0.820, 1.000}
\definecolor{p19}{rgb}{1.000, 0.810, 1.000}
\definecolor{p20}{rgb}{1.000, 0.800, 1.000}
\definecolor{p21}{rgb}{1.000, 0.790, 1.000}
\definecolor{p22}{rgb}{1.000, 0.780, 1.000}
\definecolor{p23}{rgb}{1.000, 0.770, 1.000}
\definecolor{p24}{rgb}{1.000, 0.760, 1.000}
\definecolor{p25}{rgb}{1.000, 0.750, 1.000}
\definecolor{p26}{rgb}{1.000, 0.740, 1.000}
\definecolor{p27}{rgb}{1.000, 0.730, 1.000}
\definecolor{p28}{rgb}{1.000, 0.720, 1.000}
\definecolor{p29}{rgb}{1.000, 0.710, 1.000}
\definecolor{p30}{rgb}{1.000, 0.700, 1.000}
\definecolor{p31}{rgb}{1.000, 0.690, 1.000}
\definecolor{p32}{rgb}{1.000, 0.680, 1.000}
\definecolor{p33}{rgb}{1.000, 0.670, 1.000}
\definecolor{p34}{rgb}{1.000, 0.660, 1.000}
\definecolor{p35}{rgb}{1.000, 0.650, 1.000}
\definecolor{p36}{rgb}{1.000, 0.640, 1.000}
\definecolor{p37}{rgb}{1.000, 0.630, 1.000}
\definecolor{p38}{rgb}{1.000, 0.620, 1.000}
\definecolor{p39}{rgb}{1.000, 0.610, 1.000}
\definecolor{p40}{rgb}{1.000, 0.600, 1.000}
\definecolor{p41}{rgb}{1.000, 0.590, 1.000}
\definecolor{p42}{rgb}{1.000, 0.580, 1.000}
\definecolor{p43}{rgb}{1.000, 0.570, 1.000}
\definecolor{p44}{rgb}{1.000, 0.560, 1.000}
\definecolor{p45}{rgb}{1.000, 0.550, 1.000}
\definecolor{p46}{rgb}{1.000, 0.540, 1.000}
\definecolor{p47}{rgb}{1.000, 0.530, 1.000}
\definecolor{p48}{rgb}{1.000, 0.520, 1.000}
\definecolor{p49}{rgb}{1.000, 0.510, 1.000}
\definecolor{p50}{rgb}{1.000, 0.500, 1.000}
\definecolor{p51}{rgb}{1.000, 0.490, 1.000}
\definecolor{p52}{rgb}{1.000, 0.480, 1.000}
\definecolor{p53}{rgb}{1.000, 0.470, 1.000}
\definecolor{p54}{rgb}{1.000, 0.460, 1.000}
\definecolor{p55}{rgb}{1.000, 0.450, 1.000}
\definecolor{p56}{rgb}{1.000, 0.440, 1.000}
\definecolor{p57}{rgb}{1.000, 0.430, 1.000}
\definecolor{p58}{rgb}{1.000, 0.420, 1.000}
\definecolor{p59}{rgb}{1.000, 0.410, 1.000}
\definecolor{p60}{rgb}{1.000, 0.400, 1.000}
\definecolor{p61}{rgb}{1.000, 0.390, 1.000}
\definecolor{p62}{rgb}{1.000, 0.380, 1.000}
\definecolor{p63}{rgb}{1.000, 0.370, 1.000}
\definecolor{p64}{rgb}{1.000, 0.360, 1.000}
\definecolor{p65}{rgb}{1.000, 0.350, 1.000}
\definecolor{p66}{rgb}{1.000, 0.340, 1.000}
\definecolor{p67}{rgb}{1.000, 0.330, 1.000}
\definecolor{p68}{rgb}{1.000, 0.320, 1.000}
\definecolor{p69}{rgb}{1.000, 0.310, 1.000}
\definecolor{p70}{rgb}{1.000, 0.300, 1.000}
\definecolor{p71}{rgb}{1.000, 0.290, 1.000}
\definecolor{p72}{rgb}{1.000, 0.280, 1.000}
\definecolor{p73}{rgb}{1.000, 0.270, 1.000}
\definecolor{p74}{rgb}{1.000, 0.260, 1.000}
\definecolor{p75}{rgb}{1.000, 0.250, 1.000}
\definecolor{p76}{rgb}{1.000, 0.240, 1.000}
\definecolor{p77}{rgb}{1.000, 0.230, 1.000}
\definecolor{p78}{rgb}{1.000, 0.220, 1.000}
\definecolor{p79}{rgb}{1.000, 0.210, 1.000}
\definecolor{p80}{rgb}{1.000, 0.200, 1.000}
\definecolor{p81}{rgb}{1.000, 0.190, 1.000}
\definecolor{p82}{rgb}{1.000, 0.180, 1.000}
\definecolor{p83}{rgb}{1.000, 0.170, 1.000}
\definecolor{p84}{rgb}{1.000, 0.160, 1.000}
\definecolor{p85}{rgb}{1.000, 0.150, 1.000}
\definecolor{p86}{rgb}{1.000, 0.140, 1.000}
\definecolor{p87}{rgb}{1.000, 0.130, 1.000}
\definecolor{p88}{rgb}{1.000, 0.120, 1.000}
\definecolor{p89}{rgb}{1.000, 0.110, 1.000}
\definecolor{p90}{rgb}{1.000, 0.100, 1.000}
\definecolor{p91}{rgb}{1.000, 0.090, 1.000}
\definecolor{p92}{rgb}{1.000, 0.080, 1.000}
\definecolor{p93}{rgb}{1.000, 0.070, 1.000}
\definecolor{p94}{rgb}{1.000, 0.060, 1.000}
\definecolor{p95}{rgb}{1.000, 0.050, 1.000}
\definecolor{p96}{rgb}{1.000, 0.040, 1.000}
\definecolor{p97}{rgb}{1.000, 0.030, 1.000}
\definecolor{p98}{rgb}{1.000, 0.020, 1.000}
\definecolor{p99}{rgb}{1.000, 0.010, 1.000}
\definecolor{p100}{rgb}{1.000, 0.000, 1.000}

\definecolor{v0}{rgb}{1.000, 1.000, 1.000}
 \definecolor{v1}{rgb}{0.990, 0.990, 1.000}
 \definecolor{v2}{rgb}{0.980, 0.980, 1.000}
 \definecolor{v3}{rgb}{0.970, 0.970, 1.000}
 \definecolor{v4}{rgb}{0.960, 0.960, 1.000}
 \definecolor{v5}{rgb}{0.950, 0.950, 1.000}
 \definecolor{v6}{rgb}{0.940, 0.940, 1.000}
 \definecolor{v7}{rgb}{0.930, 0.930, 1.000}
 \definecolor{v8}{rgb}{0.920, 0.920, 1.000}
 \definecolor{v9}{rgb}{0.910, 0.910, 1.000}
\definecolor{v10}{rgb}{0.900, 0.900, 1.000}
\definecolor{v11}{rgb}{0.890, 0.890, 1.000}
\definecolor{v12}{rgb}{0.880, 0.880, 1.000}
\definecolor{v13}{rgb}{0.870, 0.870, 1.000}
\definecolor{v14}{rgb}{0.860, 0.860, 1.000}
\definecolor{v15}{rgb}{0.850, 0.850, 1.000}
\definecolor{v16}{rgb}{0.840, 0.840, 1.000}
\definecolor{v17}{rgb}{0.830, 0.830, 1.000}
\definecolor{v18}{rgb}{0.820, 0.820, 1.000}
\definecolor{v19}{rgb}{0.810, 0.810, 1.000}
\definecolor{v20}{rgb}{0.800, 0.800, 1.000}
\definecolor{v21}{rgb}{0.790, 0.790, 1.000}
\definecolor{v22}{rgb}{0.780, 0.780, 1.000}
\definecolor{v23}{rgb}{0.770, 0.770, 1.000}
\definecolor{v24}{rgb}{0.760, 0.760, 1.000}
\definecolor{v25}{rgb}{0.750, 0.750, 1.000}
\definecolor{v26}{rgb}{0.740, 0.740, 1.000}
\definecolor{v27}{rgb}{0.730, 0.730, 1.000}
\definecolor{v28}{rgb}{0.720, 0.720, 1.000}
\definecolor{v29}{rgb}{0.710, 0.710, 1.000}
\definecolor{v30}{rgb}{0.700, 0.700, 1.000}
\definecolor{v31}{rgb}{0.690, 0.690, 1.000}
\definecolor{v32}{rgb}{0.680, 0.680, 1.000}
\definecolor{v33}{rgb}{0.670, 0.670, 1.000}
\definecolor{v34}{rgb}{0.660, 0.660, 1.000}
\definecolor{v35}{rgb}{0.650, 0.650, 1.000}
\definecolor{v36}{rgb}{0.640, 0.640, 1.000}
\definecolor{v37}{rgb}{0.630, 0.630, 1.000}
\definecolor{v38}{rgb}{0.620, 0.620, 1.000}
\definecolor{v39}{rgb}{0.610, 0.610, 1.000}
\definecolor{v40}{rgb}{0.600, 0.600, 1.000}
\definecolor{v41}{rgb}{0.590, 0.590, 1.000}
\definecolor{v42}{rgb}{0.580, 0.580, 1.000}
\definecolor{v43}{rgb}{0.570, 0.570, 1.000}
\definecolor{v44}{rgb}{0.560, 0.560, 1.000}
\definecolor{v45}{rgb}{0.550, 0.550, 1.000}
\definecolor{v46}{rgb}{0.540, 0.540, 1.000}
\definecolor{v47}{rgb}{0.530, 0.530, 1.000}
\definecolor{v48}{rgb}{0.520, 0.520, 1.000}
\definecolor{v49}{rgb}{0.510, 0.510, 1.000}
\definecolor{v50}{rgb}{0.500, 0.500, 1.000}
\definecolor{v51}{rgb}{0.490, 0.490, 1.000}
\definecolor{v52}{rgb}{0.480, 0.480, 1.000}
\definecolor{v53}{rgb}{0.470, 0.470, 1.000}
\definecolor{v54}{rgb}{0.460, 0.460, 1.000}
\definecolor{v55}{rgb}{0.450, 0.450, 1.000}
\definecolor{v56}{rgb}{0.440, 0.440, 1.000}
\definecolor{v57}{rgb}{0.430, 0.430, 1.000}
\definecolor{v58}{rgb}{0.420, 0.420, 1.000}
\definecolor{v59}{rgb}{0.410, 0.410, 1.000}
\definecolor{v60}{rgb}{0.400, 0.400, 1.000}
\definecolor{v61}{rgb}{0.390, 0.390, 1.000}
\definecolor{v62}{rgb}{0.380, 0.380, 1.000}
\definecolor{v63}{rgb}{0.370, 0.370, 1.000}
\definecolor{v64}{rgb}{0.360, 0.360, 1.000}
\definecolor{v65}{rgb}{0.350, 0.350, 1.000}
\definecolor{v66}{rgb}{0.340, 0.340, 1.000}
\definecolor{v67}{rgb}{0.330, 0.330, 1.000}
\definecolor{v68}{rgb}{0.320, 0.320, 1.000}
\definecolor{v69}{rgb}{0.310, 0.310, 1.000}
\definecolor{v70}{rgb}{0.300, 0.300, 1.000}
\definecolor{v71}{rgb}{0.290, 0.290, 1.000}
\definecolor{v72}{rgb}{0.280, 0.280, 1.000}
\definecolor{v73}{rgb}{0.270, 0.270, 1.000}
\definecolor{v74}{rgb}{0.260, 0.260, 1.000}
\definecolor{v75}{rgb}{0.250, 0.250, 1.000}
\definecolor{v76}{rgb}{0.240, 0.240, 1.000}
\definecolor{v77}{rgb}{0.230, 0.230, 1.000}
\definecolor{v78}{rgb}{0.220, 0.220, 1.000}
\definecolor{v79}{rgb}{0.210, 0.210, 1.000}
\definecolor{v80}{rgb}{0.200, 0.200, 1.000}
\definecolor{v81}{rgb}{0.190, 0.190, 1.000}
\definecolor{v82}{rgb}{0.180, 0.180, 1.000}
\definecolor{v83}{rgb}{0.170, 0.170, 1.000}
\definecolor{v84}{rgb}{0.160, 0.160, 1.000}
\definecolor{v85}{rgb}{0.150, 0.150, 1.000}
\definecolor{v86}{rgb}{0.140, 0.140, 1.000}
\definecolor{v87}{rgb}{0.130, 0.130, 1.000}
\definecolor{v88}{rgb}{0.120, 0.120, 1.000}
\definecolor{v89}{rgb}{0.110, 0.110, 1.000}
\definecolor{v90}{rgb}{0.100, 0.100, 1.000}
\definecolor{v91}{rgb}{0.090, 0.090, 1.000}
\definecolor{v92}{rgb}{0.080, 0.080, 1.000}
\definecolor{v93}{rgb}{0.070, 0.070, 1.000}
\definecolor{v94}{rgb}{0.060, 0.060, 1.000}
\definecolor{v95}{rgb}{0.050, 0.050, 1.000}
\definecolor{v96}{rgb}{0.040, 0.040, 1.000}
\definecolor{v97}{rgb}{0.030, 0.030, 1.000}
\definecolor{v98}{rgb}{0.020, 0.020, 1.000}
\definecolor{v99}{rgb}{0.010, 0.010, 1.000}
\definecolor{v100}{rgb}{0.000, 0.000, 1.000}

\definecolor{g0}{rgb}{1.000, 1.000, 1.000}
\definecolor{g1}{rgb}{0.990, 1.000, 0.990}
\definecolor{g2}{rgb}{0.980, 1.000, 0.980}
\definecolor{g3}{rgb}{0.970, 1.000, 0.970}
\definecolor{g4}{rgb}{0.960, 1.000, 0.960}
\definecolor{g5}{rgb}{0.950, 1.000, 0.950}
\definecolor{g6}{rgb}{0.940, 1.000, 0.940}
\definecolor{g7}{rgb}{0.930, 1.000, 0.930}
\definecolor{g8}{rgb}{0.920, 1.000, 0.920}
\definecolor{g9}{rgb}{0.910, 1.000, 0.910}
\definecolor{g10}{rgb}{0.900, 1.000, 0.900}
\definecolor{g11}{rgb}{0.890, 1.000, 0.890}
\definecolor{g12}{rgb}{0.880, 1.000, 0.880}
\definecolor{g13}{rgb}{0.870, 1.000, 0.870}
\definecolor{g14}{rgb}{0.860, 1.000, 0.860}
\definecolor{g15}{rgb}{0.850, 1.000, 0.850}
\definecolor{g16}{rgb}{0.840, 1.000, 0.840}
\definecolor{g17}{rgb}{0.830, 1.000, 0.830}
\definecolor{g18}{rgb}{0.820, 1.000, 0.820}
\definecolor{g19}{rgb}{0.810, 1.000, 0.810}
\definecolor{g20}{rgb}{0.800, 1.000, 0.800}
\definecolor{g21}{rgb}{0.790, 1.000, 0.790}
\definecolor{g22}{rgb}{0.780, 1.000, 0.780}
\definecolor{g23}{rgb}{0.770, 1.000, 0.770}
\definecolor{g24}{rgb}{0.760, 1.000, 0.760}
\definecolor{g25}{rgb}{0.750, 1.000, 0.750}
\definecolor{g26}{rgb}{0.740, 1.000, 0.740}
\definecolor{g27}{rgb}{0.730, 1.000, 0.730}
\definecolor{g28}{rgb}{0.720, 1.000, 0.720}
\definecolor{g29}{rgb}{0.710, 1.000, 0.710}
\definecolor{g30}{rgb}{0.700, 1.000, 0.700}
\definecolor{g31}{rgb}{0.690, 1.000, 0.690}
\definecolor{g32}{rgb}{0.680, 1.000, 0.680}
\definecolor{g33}{rgb}{0.670, 1.000, 0.670}
\definecolor{g34}{rgb}{0.660, 1.000, 0.660}
\definecolor{g35}{rgb}{0.650, 1.000, 0.650}
\definecolor{g36}{rgb}{0.640, 1.000, 0.640}
\definecolor{g37}{rgb}{0.630, 1.000, 0.630}
\definecolor{g38}{rgb}{0.620, 1.000, 0.620}
\definecolor{g39}{rgb}{0.610, 1.000, 0.610}
\definecolor{g40}{rgb}{0.600, 1.000, 0.600}
\definecolor{g41}{rgb}{0.590, 1.000, 0.590}
\definecolor{g42}{rgb}{0.580, 1.000, 0.580}
\definecolor{g43}{rgb}{0.570, 1.000, 0.570}
\definecolor{g44}{rgb}{0.560, 1.000, 0.560}
\definecolor{g45}{rgb}{0.550, 1.000, 0.550}
\definecolor{g46}{rgb}{0.540, 1.000, 0.540}
\definecolor{g47}{rgb}{0.530, 1.000, 0.530}
\definecolor{g48}{rgb}{0.520, 1.000, 0.520}
\definecolor{g49}{rgb}{0.510, 1.000, 0.510}
\definecolor{g50}{rgb}{0.500, 1.000, 0.500}
\definecolor{g51}{rgb}{0.490, 1.000, 0.490}
\definecolor{g52}{rgb}{0.480, 1.000, 0.480}
\definecolor{g53}{rgb}{0.470, 1.000, 0.470}
\definecolor{g54}{rgb}{0.460, 1.000, 0.460}
\definecolor{g55}{rgb}{0.450, 1.000, 0.450}
\definecolor{g56}{rgb}{0.440, 1.000, 0.440}
\definecolor{g57}{rgb}{0.430, 1.000, 0.430}
\definecolor{g58}{rgb}{0.420, 1.000, 0.420}
\definecolor{g59}{rgb}{0.410, 1.000, 0.410}
\definecolor{g60}{rgb}{0.400, 1.000, 0.400}
\definecolor{g61}{rgb}{0.390, 1.000, 0.390}
\definecolor{g62}{rgb}{0.380, 1.000, 0.380}
\definecolor{g63}{rgb}{0.370, 1.000, 0.370}
\definecolor{g64}{rgb}{0.360, 1.000, 0.360}
\definecolor{g65}{rgb}{0.350, 1.000, 0.350}
\definecolor{g66}{rgb}{0.340, 1.000, 0.340}
\definecolor{g67}{rgb}{0.330, 1.000, 0.330}
\definecolor{g68}{rgb}{0.320, 1.000, 0.320}
\definecolor{g69}{rgb}{0.310, 1.000, 0.310}
\definecolor{g70}{rgb}{0.300, 1.000, 0.300}
\definecolor{g71}{rgb}{0.290, 1.000, 0.290}
\definecolor{g72}{rgb}{0.280, 1.000, 0.280}
\definecolor{g73}{rgb}{0.270, 1.000, 0.270}
\definecolor{g74}{rgb}{0.260, 1.000, 0.260}
\definecolor{g75}{rgb}{0.250, 1.000, 0.250}
\definecolor{g76}{rgb}{0.240, 1.000, 0.240}
\definecolor{g77}{rgb}{0.230, 1.000, 0.230}
\definecolor{g78}{rgb}{0.220, 1.000, 0.220}
\definecolor{g79}{rgb}{0.210, 1.000, 0.210}
\definecolor{g80}{rgb}{0.200, 1.000, 0.200}
\definecolor{g81}{rgb}{0.190, 1.000, 0.190}
\definecolor{g82}{rgb}{0.180, 1.000, 0.180}
\definecolor{g83}{rgb}{0.170, 1.000, 0.170}
\definecolor{g84}{rgb}{0.160, 1.000, 0.160}
\definecolor{g85}{rgb}{0.150, 1.000, 0.150}
\definecolor{g86}{rgb}{0.140, 1.000, 0.140}
\definecolor{g87}{rgb}{0.130, 1.000, 0.130}
\definecolor{g88}{rgb}{0.120, 1.000, 0.120}
\definecolor{g89}{rgb}{0.110, 1.000, 0.110}
\definecolor{g90}{rgb}{0.100, 1.000, 0.100}
\definecolor{g91}{rgb}{0.090, 1.000, 0.090}
\definecolor{g92}{rgb}{0.080, 1.000, 0.080}
\definecolor{g93}{rgb}{0.070, 1.000, 0.070}
\definecolor{g94}{rgb}{0.060, 1.000, 0.060}
\definecolor{g95}{rgb}{0.050, 1.000, 0.050}
\definecolor{g96}{rgb}{0.040, 1.000, 0.040}
\definecolor{g97}{rgb}{0.030, 1.000, 0.030}
\definecolor{g98}{rgb}{0.020, 1.000, 0.020}
\definecolor{g99}{rgb}{0.010, 1.000, 0.010}
\definecolor{g100}{rgb}{0.000, 1.000, 0.000}

\definecolor{r0}{rgb}{1.000, 1.000, 1.000}
\definecolor{r1}{rgb}{1.000, 0.990, 0.990}
\definecolor{r2}{rgb}{1.000, 0.980, 0.980}
\definecolor{r3}{rgb}{1.000, 0.970, 0.970}
\definecolor{r4}{rgb}{1.000, 0.960, 0.960}
\definecolor{r5}{rgb}{1.000, 0.950, 0.950}
\definecolor{r6}{rgb}{1.000, 0.940, 0.940}
\definecolor{r7}{rgb}{1.000, 0.930, 0.930}
\definecolor{r8}{rgb}{1.000, 0.920, 0.920}
\definecolor{r9}{rgb}{1.000, 0.910, 0.910}
\definecolor{r10}{rgb}{1.000, 0.900, 0.900}
\definecolor{r11}{rgb}{1.000, 0.890, 0.890}
\definecolor{r12}{rgb}{1.000, 0.880, 0.880}
\definecolor{r13}{rgb}{1.000, 0.870, 0.870}
\definecolor{r14}{rgb}{1.000, 0.860, 0.860}
\definecolor{r15}{rgb}{1.000, 0.850, 0.850}
\definecolor{r16}{rgb}{1.000, 0.840, 0.840}
\definecolor{r17}{rgb}{1.000, 0.830, 0.830}
\definecolor{r18}{rgb}{1.000, 0.820, 0.820}
\definecolor{r19}{rgb}{1.000, 0.810, 0.810}
\definecolor{r20}{rgb}{1.000, 0.800, 0.800}
\definecolor{r21}{rgb}{1.000, 0.790, 0.790}
\definecolor{r22}{rgb}{1.000, 0.780, 0.780}
\definecolor{r23}{rgb}{1.000, 0.770, 0.770}
\definecolor{r24}{rgb}{1.000, 0.760, 0.760}
\definecolor{r25}{rgb}{1.000, 0.750, 0.750}
\definecolor{r26}{rgb}{1.000, 0.740, 0.740}
\definecolor{r27}{rgb}{1.000, 0.730, 0.730}
\definecolor{r28}{rgb}{1.000, 0.720, 0.720}
\definecolor{r29}{rgb}{1.000, 0.710, 0.710}
\definecolor{r30}{rgb}{1.000, 0.700, 0.700}
\definecolor{r31}{rgb}{1.000, 0.690, 0.690}
\definecolor{r32}{rgb}{1.000, 0.680, 0.680}
\definecolor{r33}{rgb}{1.000, 0.670, 0.670}
\definecolor{r34}{rgb}{1.000, 0.660, 0.660}
\definecolor{r35}{rgb}{1.000, 0.650, 0.650}
\definecolor{r36}{rgb}{1.000, 0.640, 0.640}
\definecolor{r37}{rgb}{1.000, 0.630, 0.630}
\definecolor{r38}{rgb}{1.000, 0.620, 0.620}
\definecolor{r39}{rgb}{1.000, 0.610, 0.610}
\definecolor{r40}{rgb}{1.000, 0.600, 0.600}
\definecolor{r41}{rgb}{1.000, 0.590, 0.590}
\definecolor{r42}{rgb}{1.000, 0.580, 0.580}
\definecolor{r43}{rgb}{1.000, 0.570, 0.570}
\definecolor{r44}{rgb}{1.000, 0.560, 0.560}
\definecolor{r45}{rgb}{1.000, 0.550, 0.550}
\definecolor{r46}{rgb}{1.000, 0.540, 0.540}
\definecolor{r47}{rgb}{1.000, 0.530, 0.530}
\definecolor{r48}{rgb}{1.000, 0.520, 0.520}
\definecolor{r49}{rgb}{1.000, 0.510, 0.510}
\definecolor{r50}{rgb}{1.000, 0.500, 0.500}
\definecolor{r51}{rgb}{1.000, 0.490, 0.490}
\definecolor{r52}{rgb}{1.000, 0.480, 0.480}
\definecolor{r53}{rgb}{1.000, 0.470, 0.470}
\definecolor{r54}{rgb}{1.000, 0.460, 0.460}
\definecolor{r55}{rgb}{1.000, 0.450, 0.450}
\definecolor{r56}{rgb}{1.000, 0.440, 0.440}
\definecolor{r57}{rgb}{1.000, 0.430, 0.430}
\definecolor{r58}{rgb}{1.000, 0.420, 0.420}
\definecolor{r59}{rgb}{1.000, 0.410, 0.410}
\definecolor{r60}{rgb}{1.000, 0.400, 0.400}
\definecolor{r61}{rgb}{1.000, 0.390, 0.390}
\definecolor{r62}{rgb}{1.000, 0.380, 0.380}
\definecolor{r63}{rgb}{1.000, 0.370, 0.370}
\definecolor{r64}{rgb}{1.000, 0.360, 0.360}
\definecolor{r65}{rgb}{1.000, 0.350, 0.350}
\definecolor{r66}{rgb}{1.000, 0.340, 0.340}
\definecolor{r67}{rgb}{1.000, 0.330, 0.330}
\definecolor{r68}{rgb}{1.000, 0.320, 0.320}
\definecolor{r69}{rgb}{1.000, 0.310, 0.310}
\definecolor{r70}{rgb}{1.000, 0.300, 0.300}
\definecolor{r71}{rgb}{1.000, 0.290, 0.290}
\definecolor{r72}{rgb}{1.000, 0.280, 0.280}
\definecolor{r73}{rgb}{1.000, 0.270, 0.270}
\definecolor{r74}{rgb}{1.000, 0.260, 0.260}
\definecolor{r75}{rgb}{1.000, 0.250, 0.250}
\definecolor{r76}{rgb}{1.000, 0.240, 0.240}
\definecolor{r77}{rgb}{1.000, 0.230, 0.230}
\definecolor{r78}{rgb}{1.000, 0.220, 0.220}
\definecolor{r79}{rgb}{1.000, 0.210, 0.210}
\definecolor{r80}{rgb}{1.000, 0.200, 0.200}
\definecolor{r81}{rgb}{1.000, 0.190, 0.190}
\definecolor{r82}{rgb}{1.000, 0.180, 0.180}
\definecolor{r83}{rgb}{1.000, 0.170, 0.170}
\definecolor{r84}{rgb}{1.000, 0.160, 0.160}
\definecolor{r85}{rgb}{1.000, 0.150, 0.150}
\definecolor{r86}{rgb}{1.000, 0.140, 0.140}
\definecolor{r87}{rgb}{1.000, 0.130, 0.130}
\definecolor{r88}{rgb}{1.000, 0.120, 0.120}
\definecolor{r89}{rgb}{1.000, 0.110, 0.110}
\definecolor{r90}{rgb}{1.000, 0.100, 0.100}
\definecolor{r91}{rgb}{1.000, 0.090, 0.090}
\definecolor{r92}{rgb}{1.000, 0.080, 0.080}
\definecolor{r93}{rgb}{1.000, 0.070, 0.070}
\definecolor{r94}{rgb}{1.000, 0.060, 0.060}
\definecolor{r95}{rgb}{1.000, 0.050, 0.050}
\definecolor{r96}{rgb}{1.000, 0.040, 0.040}
\definecolor{r97}{rgb}{1.000, 0.030, 0.030}
\definecolor{r98}{rgb}{1.000, 0.020, 0.020}
\definecolor{r99}{rgb}{1.000, 0.010, 0.010}
\definecolor{r100}{rgb}{1.000, 0.000, 0.000}

\definecolor{y0}{rgb}{1.000, 1.000, 1.000}
\definecolor{y1}{rgb}{1.000, 1.000, 0.990}
\definecolor{y2}{rgb}{1.000, 1.000, 0.980}
\definecolor{y3}{rgb}{1.000, 1.000, 0.970}
\definecolor{y4}{rgb}{1.000, 1.000, 0.960}
\definecolor{y5}{rgb}{1.000, 1.000, 0.950}
\definecolor{y6}{rgb}{1.000, 1.000, 0.940}
\definecolor{y7}{rgb}{1.000, 1.000, 0.930}
\definecolor{y8}{rgb}{1.000, 1.000, 0.920}
\definecolor{y9}{rgb}{1.000, 1.000, 0.910}
\definecolor{y10}{rgb}{1.000, 1.000, 0.900}
\definecolor{y11}{rgb}{1.000, 1.000, 0.890}
\definecolor{y12}{rgb}{1.000, 1.000, 0.880}
\definecolor{y13}{rgb}{1.000, 1.000, 0.870}
\definecolor{y14}{rgb}{1.000, 1.000, 0.860}
\definecolor{y15}{rgb}{1.000, 1.000, 0.850}
\definecolor{y16}{rgb}{1.000, 1.000, 0.840}
\definecolor{y17}{rgb}{1.000, 1.000, 0.830}
\definecolor{y18}{rgb}{1.000, 1.000, 0.820}
\definecolor{y19}{rgb}{1.000, 1.000, 0.810}
\definecolor{y20}{rgb}{1.000, 1.000, 0.800}
\definecolor{y21}{rgb}{1.000, 1.000, 0.790}
\definecolor{y22}{rgb}{1.000, 1.000, 0.780}
\definecolor{y23}{rgb}{1.000, 1.000, 0.770}
\definecolor{y24}{rgb}{1.000, 1.000, 0.760}
\definecolor{y25}{rgb}{1.000, 1.000, 0.750}
\definecolor{y26}{rgb}{1.000, 1.000, 0.740}
\definecolor{y27}{rgb}{1.000, 1.000, 0.730}
\definecolor{y28}{rgb}{1.000, 1.000, 0.720}
\definecolor{y29}{rgb}{1.000, 1.000, 0.710}
\definecolor{y30}{rgb}{1.000, 1.000, 0.700}
\definecolor{y31}{rgb}{1.000, 1.000, 0.690}
\definecolor{y32}{rgb}{1.000, 1.000, 0.680}
\definecolor{y33}{rgb}{1.000, 1.000, 0.670}
\definecolor{y34}{rgb}{1.000, 1.000, 0.660}
\definecolor{y35}{rgb}{1.000, 1.000, 0.650}
\definecolor{y36}{rgb}{1.000, 1.000, 0.640}
\definecolor{y37}{rgb}{1.000, 1.000, 0.630}
\definecolor{y38}{rgb}{1.000, 1.000, 0.620}
\definecolor{y39}{rgb}{1.000, 1.000, 0.610}
\definecolor{y40}{rgb}{1.000, 1.000, 0.600}
\definecolor{y41}{rgb}{1.000, 1.000, 0.590}
\definecolor{y42}{rgb}{1.000, 1.000, 0.580}
\definecolor{y43}{rgb}{1.000, 1.000, 0.570}
\definecolor{y44}{rgb}{1.000, 1.000, 0.560}
\definecolor{y45}{rgb}{1.000, 1.000, 0.550}
\definecolor{y46}{rgb}{1.000, 1.000, 0.540}
\definecolor{y47}{rgb}{1.000, 1.000, 0.530}
\definecolor{y48}{rgb}{1.000, 1.000, 0.520}
\definecolor{y49}{rgb}{1.000, 1.000, 0.510}
\definecolor{y50}{rgb}{1.000, 1.000, 0.500}
\definecolor{y51}{rgb}{1.000, 1.000, 0.490}
\definecolor{y52}{rgb}{1.000, 1.000, 0.480}
\definecolor{y53}{rgb}{1.000, 1.000, 0.470}
\definecolor{y54}{rgb}{1.000, 1.000, 0.460}
\definecolor{y55}{rgb}{1.000, 1.000, 0.450}
\definecolor{y56}{rgb}{1.000, 1.000, 0.440}
\definecolor{y57}{rgb}{1.000, 1.000, 0.430}
\definecolor{y58}{rgb}{1.000, 1.000, 0.420}
\definecolor{y59}{rgb}{1.000, 1.000, 0.410}
\definecolor{y60}{rgb}{1.000, 1.000, 0.400}
\definecolor{y61}{rgb}{1.000, 1.000, 0.390}
\definecolor{y62}{rgb}{1.000, 1.000, 0.380}
\definecolor{y63}{rgb}{1.000, 1.000, 0.370}
\definecolor{y64}{rgb}{1.000, 1.000, 0.360}
\definecolor{y65}{rgb}{1.000, 1.000, 0.350}
\definecolor{y66}{rgb}{1.000, 1.000, 0.340}
\definecolor{y67}{rgb}{1.000, 1.000, 0.330}
\definecolor{y68}{rgb}{1.000, 1.000, 0.320}
\definecolor{y69}{rgb}{1.000, 1.000, 0.310}
\definecolor{y70}{rgb}{1.000, 1.000, 0.300}
\definecolor{y71}{rgb}{1.000, 1.000, 0.290}
\definecolor{y72}{rgb}{1.000, 1.000, 0.280}
\definecolor{y73}{rgb}{1.000, 1.000, 0.270}
\definecolor{y74}{rgb}{1.000, 1.000, 0.260}
\definecolor{y75}{rgb}{1.000, 1.000, 0.250}
\definecolor{y76}{rgb}{1.000, 1.000, 0.240}
\definecolor{y77}{rgb}{1.000, 1.000, 0.230}
\definecolor{y78}{rgb}{1.000, 1.000, 0.220}
\definecolor{y79}{rgb}{1.000, 1.000, 0.210}
\definecolor{y80}{rgb}{1.000, 1.000, 0.200}
\definecolor{y81}{rgb}{1.000, 1.000, 0.190}
\definecolor{y82}{rgb}{1.000, 1.000, 0.180}
\definecolor{y83}{rgb}{1.000, 1.000, 0.170}
\definecolor{y84}{rgb}{1.000, 1.000, 0.160}
\definecolor{y85}{rgb}{1.000, 1.000, 0.150}
\definecolor{y86}{rgb}{1.000, 1.000, 0.140}
\definecolor{y87}{rgb}{1.000, 1.000, 0.130}
\definecolor{y88}{rgb}{1.000, 1.000, 0.120}
\definecolor{y89}{rgb}{1.000, 1.000, 0.110}
\definecolor{y90}{rgb}{1.000, 1.000, 0.100}
\definecolor{y91}{rgb}{1.000, 1.000, 0.090}
\definecolor{y92}{rgb}{1.000, 1.000, 0.080}
\definecolor{y93}{rgb}{1.000, 1.000, 0.070}
\definecolor{y94}{rgb}{1.000, 1.000, 0.060}
\definecolor{y95}{rgb}{1.000, 1.000, 0.050}
\definecolor{y96}{rgb}{1.000, 1.000, 0.040}
\definecolor{y97}{rgb}{1.000, 1.000, 0.030}
\definecolor{y98}{rgb}{1.000, 1.000, 0.020}
\definecolor{y99}{rgb}{1.000, 1.000, 0.010}
\definecolor{y100}{rgb}{1.000, 1.000, 0.000}

\definecolor{w0}{rgb}{1.000, 1.000, 1.000}
\definecolor{w1}{rgb}{0.990, 0.990, 0.990}
\definecolor{w2}{rgb}{0.980, 0.980, 0.980}
\definecolor{w3}{rgb}{0.970, 0.970, 0.970}
\definecolor{w4}{rgb}{0.960, 0.960, 0.960}
\definecolor{w5}{rgb}{0.950, 0.950, 0.950}
\definecolor{w6}{rgb}{0.940, 0.940, 0.940}
\definecolor{w7}{rgb}{0.930, 0.930, 0.930}
\definecolor{w8}{rgb}{0.920, 0.920, 0.920}
\definecolor{w9}{rgb}{0.910, 0.910, 0.910}
\definecolor{w10}{rgb}{0.900, 0.900, 0.900}
\definecolor{w11}{rgb}{0.890, 0.890, 0.890}
\definecolor{w12}{rgb}{0.880, 0.880, 0.880}
\definecolor{w13}{rgb}{0.870, 0.870, 0.870}
\definecolor{w14}{rgb}{0.860, 0.860, 0.860}
\definecolor{w15}{rgb}{0.850, 0.850, 0.850}
\definecolor{w16}{rgb}{0.840, 0.840, 0.840}
\definecolor{w17}{rgb}{0.830, 0.830, 0.830}
\definecolor{w18}{rgb}{0.820, 0.820, 0.820}
\definecolor{w19}{rgb}{0.810, 0.810, 0.810}
\definecolor{w20}{rgb}{0.800, 0.800, 0.800}
\definecolor{w21}{rgb}{0.790, 0.790, 0.790}
\definecolor{w22}{rgb}{0.780, 0.780, 0.780}
\definecolor{w23}{rgb}{0.770, 0.770, 0.770}
\definecolor{w24}{rgb}{0.760, 0.760, 0.760}
\definecolor{w25}{rgb}{0.750, 0.750, 0.750}
\definecolor{w26}{rgb}{0.740, 0.740, 0.740}
\definecolor{w27}{rgb}{0.730, 0.730, 0.730}
\definecolor{w28}{rgb}{0.720, 0.720, 0.720}
\definecolor{w29}{rgb}{0.710, 0.710, 0.710}
\definecolor{w30}{rgb}{0.700, 0.700, 0.700}
\definecolor{w31}{rgb}{0.690, 0.690, 0.690}
\definecolor{w32}{rgb}{0.680, 0.680, 0.680}
\definecolor{w33}{rgb}{0.670, 0.670, 0.670}
\definecolor{w34}{rgb}{0.660, 0.660, 0.660}
\definecolor{w35}{rgb}{0.650, 0.650, 0.650}
\definecolor{w36}{rgb}{0.640, 0.640, 0.640}
\definecolor{w37}{rgb}{0.630, 0.630, 0.630}
\definecolor{w38}{rgb}{0.620, 0.620, 0.620}
\definecolor{w39}{rgb}{0.610, 0.610, 0.610}
\definecolor{w40}{rgb}{0.600, 0.600, 0.600}
\definecolor{w41}{rgb}{0.590, 0.590, 0.590}
\definecolor{w42}{rgb}{0.580, 0.580, 0.580}
\definecolor{w43}{rgb}{0.570, 0.570, 0.570}
\definecolor{w44}{rgb}{0.560, 0.560, 0.560}
\definecolor{w45}{rgb}{0.550, 0.550, 0.550}
\definecolor{w46}{rgb}{0.540, 0.540, 0.540}
\definecolor{w47}{rgb}{0.530, 0.530, 0.530}
\definecolor{w48}{rgb}{0.520, 0.520, 0.520}
\definecolor{w49}{rgb}{0.510, 0.510, 0.510}
\definecolor{w50}{rgb}{0.500, 0.500, 0.500}
\definecolor{w51}{rgb}{0.490, 0.490, 0.490}
\definecolor{w52}{rgb}{0.480, 0.480, 0.480}
\definecolor{w53}{rgb}{0.470, 0.470, 0.470}
\definecolor{w54}{rgb}{0.460, 0.460, 0.460}
\definecolor{w55}{rgb}{0.450, 0.450, 0.450}
\definecolor{w56}{rgb}{0.440, 0.440, 0.440}
\definecolor{w57}{rgb}{0.430, 0.430, 0.430}
\definecolor{w58}{rgb}{0.420, 0.420, 0.420}
\definecolor{w59}{rgb}{0.410, 0.410, 0.410}
\definecolor{w60}{rgb}{0.400, 0.400, 0.400}
\definecolor{w61}{rgb}{0.390, 0.390, 0.390}
\definecolor{w62}{rgb}{0.380, 0.380, 0.380}
\definecolor{w63}{rgb}{0.370, 0.370, 0.370}
\definecolor{w64}{rgb}{0.360, 0.360, 0.360}
\definecolor{w65}{rgb}{0.350, 0.350, 0.350}
\definecolor{w66}{rgb}{0.340, 0.340, 0.340}
\definecolor{w67}{rgb}{0.330, 0.330, 0.330}
\definecolor{w68}{rgb}{0.320, 0.320, 0.320}
\definecolor{w69}{rgb}{0.310, 0.310, 0.310}
\definecolor{w70}{rgb}{0.300, 0.300, 0.300}
\definecolor{w71}{rgb}{0.290, 0.290, 0.290}
\definecolor{w72}{rgb}{0.280, 0.280, 0.280}
\definecolor{w73}{rgb}{0.270, 0.270, 0.270}
\definecolor{w74}{rgb}{0.260, 0.260, 0.260}
\definecolor{w75}{rgb}{0.250, 0.250, 0.250}
\definecolor{w76}{rgb}{0.240, 0.240, 0.240}
\definecolor{w77}{rgb}{0.230, 0.230, 0.230}
\definecolor{w78}{rgb}{0.220, 0.220, 0.220}
\definecolor{w79}{rgb}{0.210, 0.210, 0.210}
\definecolor{w80}{rgb}{0.200, 0.200, 0.200}
\definecolor{w81}{rgb}{0.190, 0.190, 0.190}
\definecolor{w82}{rgb}{0.180, 0.180, 0.180}
\definecolor{w83}{rgb}{0.170, 0.170, 0.170}
\definecolor{w84}{rgb}{0.160, 0.160, 0.160}
\definecolor{w85}{rgb}{0.150, 0.150, 0.150}
\definecolor{w86}{rgb}{0.140, 0.140, 0.140}
\definecolor{w87}{rgb}{0.130, 0.130, 0.130}
\definecolor{w88}{rgb}{0.120, 0.120, 0.120}
\definecolor{w89}{rgb}{0.110, 0.110, 0.110}
\definecolor{w90}{rgb}{0.100, 0.100, 0.100}
\definecolor{w91}{rgb}{0.090, 0.090, 0.090}
\definecolor{w92}{rgb}{0.080, 0.080, 0.080}
\definecolor{w93}{rgb}{0.070, 0.070, 0.070}
\definecolor{w94}{rgb}{0.060, 0.060, 0.060}
\definecolor{w95}{rgb}{0.050, 0.050, 0.050}
\definecolor{w96}{rgb}{0.040, 0.040, 0.040}
\definecolor{w97}{rgb}{0.030, 0.030, 0.030}
\definecolor{w98}{rgb}{0.020, 0.020, 0.020}
\definecolor{w99}{rgb}{0.010, 0.010, 0.010}
\definecolor{w100}{rgb}{0.000, 0.000, 0.000}

\definecolor{cleanN}{rgb}{0.70,0.70,0.70}
\definecolor{brownN}{rgb}{0.80,0.00,0.00}
\definecolor{whiteN}{rgb}{0.00,0.00,0.80}
\definecolor{pinkN}{rgb}{0.80,0.00,1.00}

\definecolor{verde}{rgb}{0.00,0.00,1.00}
\definecolor{naranja}{rgb}{0.00,1.00,0.00} 

\usepackage[activate]{microtype}
\newcommand{\ie}{i.\,e.}
\newcommand{\eg}{e.\,g.}

\usepackage[natbib, bibencoding=utf8, citestyle=numeric, bibstyle=ieee, maxbibnames=999, maxcitenames=2, mincitenames=1, sortcites]{biblatex}
\bibliography{mybib}

\title{INTERSPEECH 2009 Emotion Challenge Revisited: \\Benchmarking 15 Years of Progress in Speech Emotion Recognition}
\name[affiliation={1}]{Andreas}{Triantafyllopoulos}
\name[affiliation={1}]{Anton}{Batliner}
\name[affiliation={2}]{Simon}{Rampp}
\name[affiliation={1}]{Manuel}{Milling}
\name[affiliation={1,2,3}]{Björn}{Schuller}
\address{
  $^1$CHI -- Chair of Health Informatics, MRI, Technical University of Munich, Germany \\
  $^2$EIHW -- Embedded Intelligence for Health Care \& Wellbeing, University of Augsburg, Germany \\
  $^3$GLAM -- Group on Language, Audio, \& Music, Imperial College London, UK}
\email{andreas.triantafyllopoulos@tum.de}
\keywords{Speech emotion recognition, Deep learning}
\begin{document}

\maketitle
 
\begin{abstract}
We revisit the INTERSPEECH 2009 Emotion Challenge -- the first ever speech emotion recognition (SER) challenge -- and evaluate a series of deep learning models that are representative of the major advances in SER research in the time since then.
We start by training each model using a fixed set of hyperparameters, and further fine-tune the best-performing models of that initial setup with a grid search.
Results are always reported on the official test set with a separate validation set only used for early stopping.
Most models score below or close to the official baseline, while they marginally outperform the original challenge winners after hyperparameter tuning.
Our work illustrates that, despite recent progress, FAU-AIBO remains a very challenging benchmark.
An interesting corollary is that newer methods do not consistently outperform older ones, showing that progress towards `solving' SER is not necessarily monotonic.
\end{abstract}

\section{Introduction}

Standardised benchmarks form the backbone of reproducible science and enable the research community to showcase its progress towards a common objective.
\Ac{SER} is one subfield of speech science where several benchmarks exist, mostly in the form of challenges like:
the INTERSPEECH 2009 Emotion Challenge (ComParE)~\citep{Schuller09-TI2a}, 
%BS: added_
the first official challenge on SER, 
which was followed by several iterations covering a wide gamut of paralinguistic tasks;
the Audio-Visual Emotion Challenge (AVEC)~\citep{Schuller11-ATF};
MuSe~\citep{Stappen20-M2C};
OMG~\citep{Barros18-TOE};
just recently, the Odyssey 2024 \ac{SER} challenge; and others.
%\footnote{\url{https://www.odyssey2024.org/emotion-recognition-challenge}};
% the Audio-Visual Emotion Challenge (AVEC) series which focused on audio-visual emotion recognition, as well as health-related speech tasks~\citep{Schuller11-ATF};
% just recently, the (ongoing) Odyssey 2024 \ac{SER} challenge\footnote{\url{https://www.odyssey2024.org/emotion-recognition-challenge}};
% and others. %BS: At least mention our "others" such as MuSe and MER (the Chinese series)? OMG and EmotiW are perhaps also noteworthy?

However, while such challenges form excellent `proving grounds' for the prevalent methods at a particular \emph{Zeitgeist}, they are rarely revisited when newer approaches emerge.
Furthermore, popular datasets oftentimes suffer from non-standardised folds (as in the case of IEMOCAP~\citep{Busso08-IIE}) or iterative releases (like the recent MSP-Podcast~\citep{Lotfian17-BNE}, with a new version being released almost every year), which makes it harder to obtain a consistent comparison of all different methods.

% AT: maybe discuss paper reviews in the journal version? would be too long here
% AT: but I do see it as the other altenative
% On the other hand, the alternative tool to measure progress are review papers.

In the present contribution -- and on the occasion of its 15\textsuperscript{th} anniversary -- we focus exclusively on FAU-AIBO~\citep{Batliner04-YST,Steidl09-ACO}, the dataset used for the first-ever \ac{SER} challenge in 2009.
The challenge contained two alternative formulations of emotion:
a 2-class problem, where participants had to differentiate between \emph{negative} and \emph{non-negative} emotions;
and a 5-class problem, where participants had to classify an utterance as \emph{angry} (A), \emph{neutral} (N), \emph{motherese}/\emph{joyful} (P), \emph{emphatic} (E), with a 5\textsuperscript{th} \emph{rest} (R) class.
As the challenge ran before the advent of the `\ac{DL} era', participants never benefited from these advances.
Moreover, 
% as the test data was never publicly released \at{is this really true?}
% \ab{no, they were released, and for me it is strange that they seemingly have not (never?) been used in the exact IS09 constellation. Might simply characterise science ... or mirror our results that IS09is very competitive?}, 
as newer datasets emerged over time, FAU-AIBO has been relatively overlooked.
% \at{tbh, I would skip the foloowing 2-3 sentences from you Anton, as we risk offending some reviewers who misused the data ;) Also, for the sake of fairness, we should cite some papers, and we don't have the space for that. I would save it for the journal version if you agree}
% \ab{I don't mind such strategic considerations :-) but I'm still not sure whether we simply should not say anything about that ...: when you write `relatively overlooked': it has been cited over 1200 times so people might wonder: why `overlooked'? The point is that is has evolved into an `any time we refer to the (hi)story of SER we refer to this paper because anybody else did it' paper. And I would not say they `misused' -- after the challenge, people were free to partition into whatever although of course, we would have liked them to use our partitions. And the prob was that we did not have a dev partition. maybe we could point this out as a possible reason? By that, we blame the authors of our own paper that reasonably cannot be reviewers -- although, as it is double blind ... :-) }
Specifically, although \cite{Schuller09-TI2a} has been referred to rather often -- being a standard reference to an early paper presenting spontaneous emotions, clear partitioning, and baselines -- to the best of our knowledge, in the last 15 years there has been only a limited number of studies that unequivocally used the same configuration of train and test partition with identical number of items in each class.
% SR: Should the dashes "--" above be used standalone?
% \at{I also found Zhao19-EDS which ostensibly uses the same partition as us, though they do not refer to a dev set}
% At least, in the few studies where the same partitioning is mentioned, no precise figures are given.\todo{citation?}
In fact, sometimes it is explicitly mentioned that ``the results presented in this paper are not directly
comparable with those found using the 2009-challenge data''~\citep{Cummins17-AID}.
On a positive note, this makes it a perfect test case for a long-overdue retrospective, centred on the question of whether the community has `solved' or at least substantially progressed on the problem of \ac{SER} in the intervening years.

To answer this question, we run a large-scale study of several `high-profile' advances that have emerged after 2009:
We start from larger, more comprehensive feature sets which defined the \ac{SER} landscape until ca.\ 2016, where we train both \acp{MLP} on their static and \acp{LSTM} on their dynamic versions.
%BS: added:
% as were provided by the challenge.
After that, we move on to the first end-to-end \acp{CRNN} as well as spectrogram-based \acp{CNN} benefiting from transfer learning.
Finally, we investigate the more recent transformers pre-trained with \ac{SSL}.
We note that such previous large-scale experiments have only been carried out within a particular architecture family~\citep{Fayek17-EDL, Zhao19-EDS, Wagner23-DOT}.
We additionally analyse our results with respect to inter-model agreement, examine whether hard-to-classify cases are also those where human annotators disagree the most, and try to measure whether progress is monotonic with respect to the year a model was introduced or its size (\ie, computational complexity).
We note that we exclude advances on linguistics due to space limitations.
%BS: The corpus allows for LINGUISTIC processing - a lot has happened here - see LLMs! But I think you focus on acoustics only - this seems very important, as perhaps, by adding LLMs here, you WOULD get a massive boost? Great for a next paper, e.g., with Iosif?

\section{Previous work on FAU-AIBO}

\noindent
\textbf{Challenge:}
The official challenge baseline consisted of HMM modelling of dynamic features or \ac{SVM} modelling of static features %(computed by the application of functionals on dynamic features), 
each combined with SMOTE oversampling to mitigate class imbalance and a separate standardisation applied on the training and test sets 
% (to account for systematic differences in the two recording locations)
~\citep{Schuller09-TI2a}, with static features yielding moderately better performance.
The challenge featured two winners:% for two of the three tracks: %(no participant beat the baseline in the third track):
The \emph{open performance} sub-challenge was won by \citet{Dumouchel09-CAL}, who employed \acp{GMM} trained on cepstral and expert prosodic and vocal tract features;
The \emph{classifier performance} sub-challenge was won by \citet{Lee11-ERU}, who used the official static features but with a divide-and-conquer cascade classification approach.
\citet{Kockmann09-BUO} obtained the best performance on the 5-class problem by employing a fusion of different \acp{GMM} trained on functionals (but were only $0.1\%$ better than \citet{Lee11-ERU}).
% A review of all challenge submissions in \citet{Schuller11-RRE} found an overall tendency towards smaller, carefully-designed features outperforming `brute-force' approaches.
% Furthermore, they showed that a fusion of the top approaches led to additional performance gains, indicating that the different submissions were, to some extent, complementary.
A review found an overall tendency towards smaller, carefully-designed features over `brute-force' approaches~\citep{Schuller11-RRE}, and a fusion of the top approaches led to additional performance gains~\citep{Schuller11-RRE}.

\noindent
\textbf{Beyond the challenge:}
Researchers continued to improve performance on FAU-AIBO after the end of the challenge.
Closely related to our work, \citet{Cummins17-AID} and \citet{Zhao19-EDS} both report better performance than the challenge winners using transfer-learning from \acp{CNN} pre-trained on image data; however, the former use different splits than the challenge and the latter perform a very extensive hyperparameter search (a total of $15$ hyperparameters were optimised, resulting in a search space much larger than the one we employ here).

\section{Methodology}

\subsection{Dataset}
We use the official dataset of the  INTERSPEECH 2009 Emotion Challenge~\citep{Schuller09-TI2a}, FAU-AIBO.
It is a dataset of German children's speech collected in a Wizard-of-Oz scenario and annotated on the word-level for the presence of $11$ emotional/communicative states by $5$ raters~\citep{Batliner04-YST, Steidl09-ACO}.
Subsequently, segmented words have been aggregated to meaningful chunks using manual semantic and prosodic criteria.
Accordingly, annotated states have been mapped to $2$- and $5$-class categorisation using a set of heuristics, which forms a final dataset of $18\,216$ chunks used for the challenge.
The data is heavily imbalanced towards the neutral/non-negative classes.
The data was collected from two schools, with one school set aside for testing (\emph{Mont}) and one set aside for training (\emph{Ohm}); we use the same partitioning for our experiments.
Additionally, we create a small validation set comprising the last two speakers of the training set (speakers are denoted by number IDs): \emph{Ohm\_31} and \emph{Ohm\_32}, similar to \citep{Triantafyllopoulos21-DSC}.
% Note that the data and details on (pre-)processing are extensively documented in \citep{Steidl09-ACO}.
Note that the data are extensively documented in \citep{Steidl09-ACO}.
%AB: letzten Satz damit kein reviewer kritisieren kann, dass wir die Daten nicht beschreiben. Sollte eigentlich kar sein, aber ...

% SR: seen the comment of AB, but it might make sense to at least mention that the dataset is imbalanced here, since it was already mentioned in the previous work section that the baseline used SMOTE.

\subsection{Models}
%BS: added:
The Computational Paralinguistics Challenge (ComParE) continued the first challenge for a further 14 years. 
Our selection of models is largely based on the ComParE series' 
baselines and best-performing winners of each year and on prevalent trends in the last decade of \ac{SER} research.
We briefly describe each model below, but also include an appendix with linked model states as well as plan to release the source code for our experiments\footnote{\url{github.com/ATriantafyllopoulos/is24-interspeech09-ser-revisited}}.

%SR: So the '09-'16 is recreated by using LSTMs? Maybe this is a good point to mention above
\noindent
\textbf{'09-'16: openSMILE feature sets} -- In the follow-up iterations of the ComParE Challenge, newer versions of paralinguistic features were introduced.
In general, these feature sets were larger and covered a wider gamut of acoustic and prosodic features.
However, that period saw a parallel pursuit for \emph{smaller} expert-driven feature sets~\citep{Eyben15-TGM}.
To accommodate both, we use both the static (`functionals') and dynamic (`low-level descriptors') versions of the official \textsc{IS09}-\textsc{IS13} and \textsc{IS16} feature sets, as well as the \textsc{eGeMAPS} feature set~\citep{Eyben15-TGM} as provided in the latest version of the openSMILE toolkit~\citep{Eyben10-OTM}.
For the \emph{dynamic} features, we used a $2$-layered \ac{LSTM} model with $32$ hidden units, followed by mean pooling over time, one hidden linear layer with $32$ neurons and ReLU acivation, and one output linear layer; all hidden layers are followed by a dropout of $0.5$; these models are denoted with $^d$.
Additionally, we train $3$-layered \acp{MLP} with $64$ hidden units each, a dropout of $0.5$, and ReLU activation for the \emph{static} features; these models are denoted with $^s$. % for each respective feature set.

\noindent
\textbf{'12-'23: ImageNet pre-training} -- Following the introduction of ImageNet and the first \acp{CNN} trained on it in 2012, such networks were subsequently introduced in the audio and speech domains by substituting images with (pictorial representations of) spectrograms~\citep{Cummins17-AID, Zhao19-EDS} -- a practice that is relevant to this day, with audio transformer models oftentimes initialised with states pre-trained on ImageNet~\citep{Gong21-AAS}.
%
% SR: arent spectrograms pictoral representations of signals already?
% AT: it depends on how you feed them in, as raw numbers or converted to RGB
%
We use \textsc{AlexNet}, \textsc{Resnet50}, all versions of \textsc{VGG} ($^{11,13,16,19}$), EfficientNet-B0 (\textsc{EffNet}), the tiny, small, base, and large versions of \textsc{ConvNeXt} ($^{t,b,s,l}$), and the tiny, base, and small versions of the \textsc{Swin} Transformer ($^{t,b,s}$).
In all cases, we use the best-performing model state on ImageNet as available in the \textsc{torchvision-v0.16.0} package.%(see supplementary material for exact versions).
%SR: Did you use a different version of torchvision? In the reqs 0.16.0 is pinned, just to be safe.
%AT: I think I did, but I'll double check
As features, we always used the Mel-spectrograms generated for \textsc{CNN14} (see below), \ie, $64$ Mels with a window size of $32$\,ms and a hop size of $10$\,ms; the resulting matrices were then replicated over the three dimensions to generate the $3$-channel input that is required by models designed for computer vision tasks.
%SR: maybe "designed for" computer vision

\noindent
\textbf{'16-'23: End-to-end} -- Subsequent years saw the introduction of \emph{end-to-end} models, \ie, models trained directly on raw audio input for the target task without any prior feature pre-processing. %SR: or preprocessing (rel to previous section)
% SR: to me end-to-end means more that you just throw in the data and get the results, but in my understanding prior training e.g. ImageNet weights is not exclusive here. Not sure maybe i understand the point wrong.
These models were especially successful in the case of time-continous \ac{SER}, which requires predicting the emotion of very short audio frames, and essentially follow the \ac{CRNN} architecture.
We use two particular instantiations introduced by \citet{Tzirakis18-ETE} (\textsc{CRNN}$^{18}$) and \citet{Zhao19-SER} (\textsc{CRNN}$^{19}$).

\noindent
\textbf{'16-'23: Supervised audio pre-training}
-- In parallel to ImageNet pre-training, there were also efforts to collect similar large-scale datasets for audio where networks could be pre-trained in a supervised fashion.
Two notable examples are VoxCeleb and AudioSet, both collected from YouTube, with the former targeted to speaker identification and the latter to general audio tagging.
VoxCeleb formed the basis for training speaker embedding models (\ie, `x-vectors') using \acp{TDNN}, of which we use a more recent and improved attention-based model (\textsc{ETDNN})~\citep{Desplanques20-EEC}.
AudioSet in turn inspired the use of VGG-based convolutional networks, such as \textsc{CNN14} introduced in \acp{PANN}~\citep{Kong20-PLS} which was shown to also transfer well to \ac{SER} tasks~\citep{Triantafyllopoulos21-TRO}, and later, transformer-based models such as \textsc{AST}~\citep{Gong21-AAS}.
In addition, the introduction of the \textsc{Whisper} architecture~\citep{Radford23-RSR} 
led to a renaissance of supervised training for \ac{ASR} and we thus include it in our experiments -- albeit only the three smallest available variants ($^{t,b,s}$) due to hardware constraints.

% SR: the "Radford23-RSR" cite looks different
\noindent
\textbf{'20-'23: Self-supervised audio pre-training}
-- The introduction of transformers and the advent of self-supervised pre-training for computer vision and \ac{NLP} also propagated to the speech and audio domain.
The two dominant architectures here are wav2vec2.0~\citep{Baevski20-W2V}, which includes a convolutional backend followed by a transformer decoder trained to reconstruct its own quantised intermediate representations, and HuBert~\citep{Hsu21-HSS}, a full-transformer model trained on masked token prediction.
These models have yielded significant advances in \ac{SER}~\citep{Wagner23-DOT}, which was partially accredited to their ability to simultaneously encode linguistic and paralinguistic information~\citep{Triantafyllopoulos22-PSE}.
In this work, we use the pretrained states from the \emph{base} and \emph{large} variants of wav2vec2.0 and HuBert (\textsc{W2V2}$^{b,l}$, \textsc{HUB}$^{b,l}$), a multilingual model trained on VoxPopuli~\citep{Wang21-VAL} (\textsc{W2V2}$^{m}$), a \emph{`robust'} version of wav2vec2.0 trained on more data~\citep{Hsu21-RW2} (\textsc{W2V2}$^{r}$), as well as the pruned version of that model further fine-tuned for dimensional \ac{SER} on MSP-Podcast~\citep{Wagner23-DOT} (\textsc{W2V2}$^{e}$).
Similar to \citet{Wagner23-DOT}, we add an output $2-$layered \ac{MLP} which takes the pooled hidden embeddings of the last layer as input.

\subsection{Experiments}

To constrain our space of hyperparameters, we conduct two experimental phases.
In the first \textbf{exploration phase}, we test all $43$ investigated models using a fixed set of hyperparameters.
Specifically, we use the Adam optimiser with a learning rate of $0.0001$ and a batch size of $4$ for $30$ epochs.
In the following \textbf{tuning phase}, we further optimise a larger set of hyperparameters for the $5$ best-performing models from the exploration phase, doing a grid search over optimisers $\{\text{Adam}, \text{SGD}\}$, learning rates $\{0.01,0.001,0.0001\}$, and batch sizes $\{4,8,16\}$, while training each configuration for $50$ epochs.
In total, this results in $43$ runs for the exploration phase and an additional $90$ runs for the tuning phase (for each of the two challenge tasks).
To account for variable-length sequences in training, we 
%BS: "randomly" - is this reproducible by others (e.g., giving seed number and function or github to experiments via link?)
randomly cropped/padded all chunks to a fixed length of $3$ seconds (different cropping/padding was applied on each instance across different epochs; random seed was fixed and can be reproduced) when using dynamic features (including the raw audio); during inference, we used the original utterances, only padding those shorter than $2$ seconds with silence. 
%SR: this also for spectrograms right?
%AT: yes, but conceptually it's the same (no need to say it was done for the frames)
Our loss function is the standard categorical cross-entropy, where -- in order to account for the severe class imbalance -- we further weigh the contribution of each instance by the inverse frequency of its true label on the training set, similar to \citet{Zhao19-EDS}.
Except for \textsc{W2V2} and \textsc{HUB}, where we freeze the feature extractors, all model parameters are fine-tuned.

In all cases, we use the defined validation set (comprising two speakers from the original training set) to select the best-performing epoch for each model, which we then proceed to evaluate on the test set.
Strictly speaking, this results in different training data from some challenge participants, as they typically retrained their models on the entire training set after optimising their hyperparameters (\eg, via cross-validation).
However, given the propensity of \acp{DNN} to overfit when trained too long, we found the use of such a validation set necessary for early stopping.

\section{Results \& Discussion}

\begin{table}[!t]
    \centering
    \caption{
    UAR results for all tested models in the \textbf{exploration phase} using our standard hypermarameters (Adam, $0.01$, $4$) for both the $2$- and the $5$-class model.
    % See supplementary material for a user-friendlier, interactive visualisation of this table.
    }
    \label{tab:exploration}
\begin{tabular}{lrll}
\toprule
          Model &  Year &                   $2$-class &                   $5$-class \\
\midrule
    \textsc{IS09}$^s$ &  2009 & \cellcolor{w23}{.620} &  \cellcolor{w1}{.294} \\
    \textsc{IS09}$^d$ &  2009 & \cellcolor{w48}{.670} & \cellcolor{w25}{.347} \\
    \textsc{IS10}$^s$ &  2010 & \cellcolor{w49}{.670} & \cellcolor{w25}{.348} \\
    \textsc{IS10}$^d$ &  2010 & \cellcolor{w58}{.689} & \cellcolor{w50}{.406} \\
    \textsc{IS11}$^s$ &  2011 &  \cellcolor{w0}{.499} &  \cellcolor{w0}{.201} \\
    \textsc{IS11}$^d$ &  2011 & \cellcolor{w48}{.670} & \cellcolor{w36}{.373} \\
    \textsc{IS12}$^s$ &  2012 &  \cellcolor{w0}{.501} &  \cellcolor{w0}{.201} \\
    \textsc{AlexNet} &  2012 &  \cellcolor{w0}{.503} &  \cellcolor{w0}{.200} \\
    \textsc{IS12}$^d$ &  2012 & \cellcolor{w53}{.679} & \cellcolor{w36}{.373} \\
    \textsc{IS13}$^d$ &  2013 & \cellcolor{w45}{.664} & \cellcolor{w38}{.378} \\
    \textsc{IS13}$^s$ &  2013 &  \cellcolor{w0}{.498} &  \cellcolor{w0}{.201} \\
    \textsc{eGeMAPS}$^s$ &  2015 & \cellcolor{w22}{.618} &  \cellcolor{w0}{.245} \\
    \textsc{eGeMAPS}$^d$ &  2015 & \cellcolor{w47}{.667} & \cellcolor{w46}{.397} \\
    \textsc{Resnet50} &  2015 & \cellcolor{w58}{.688} & \cellcolor{w60}{.428} \\
    \textsc{VGG}$^{19}$ &  2016 &  \cellcolor{w0}{.499} &  \cellcolor{w0}{.204} \\
    \textsc{IS16}$^s$ &  2016 &  \cellcolor{w0}{.500} &  \cellcolor{w0}{.201} \\
    \textsc{VGG}$^{16}$ &  2016 & \cellcolor{w36}{.646} & \cellcolor{w49}{.404} \\
    \textsc{IS16}$^d$ &  2016 & \cellcolor{w41}{.655} & \cellcolor{w40}{.382} \\
    \textsc{VGG}$^{13}$ &  2016 & \cellcolor{w46}{.665} & \cellcolor{w41}{.385} \\
    \textsc{VGG}$^{11}$ &  2016 & \cellcolor{w46}{.666} &  \cellcolor{w0}{.200} \\
    \textsc{EffNet} &  2019 & \cellcolor{w48}{.669} & \cellcolor{w27}{.353} \\
    \textsc{CRNN}$^{18}$ &  2018 & \cellcolor{w53}{.680} & \cellcolor{w44}{.392} \\
    \textsc{CRNN}$^{19}$ &  2019 & \cellcolor{w55}{.683} & \cellcolor{w35}{.372} \\
    \textsc{W2V2}$^l$ &  2020 &  \cellcolor{w0}{.500} &  \cellcolor{w0}{.200} \\
    \textsc{W2V2}$^b$  &  2020 &  \cellcolor{w0}{.500} &  \cellcolor{w0}{.200} \\
    \textsc{ConvNeXt}$^t$  &  2020 &  \cellcolor{w0}{.500} & \cellcolor{w48}{.400} \\
    \textsc{ConvNeXt}$^l$  &  2020 & \cellcolor{w45}{.663} & \cellcolor{w52}{.409} \\
    \textsc{ConvNeXt}$^b$  &  2020 & \cellcolor{w46}{.665} & \cellcolor{w53}{.412} \\
    \textsc{ConvNeXt}$^s$ &  2020 & \cellcolor{w50}{.674} & \cellcolor{w50}{.406} \\
    \textsc{ETDNN} &  2020 & \cellcolor{w53}{.678} & \cellcolor{w50}{.404} \\
    \textsc{CNN14} &  2020 & \cellcolor{w60}{.692} & \cellcolor{w45}{.394} \\
    \textsc{HUB}$^b$ &  2021 &  \cellcolor{w0}{.500} &  \cellcolor{w0}{.200} \\
    \textsc{Swin}$^b$ &  2021 &  \cellcolor{w0}{.528} &  \cellcolor{w0}{.200} \\
    \textsc{Swin}$^t$ &  2021 &  \cellcolor{w0}{.530} &  \cellcolor{w0}{.242} \\
    \textsc{AST} &  2021 &  \cellcolor{w0}{.535} &  \cellcolor{w4}{.300} \\
    \textsc{W2V2}$^m$ &  2021 & \cellcolor{w33}{.640} & \cellcolor{w48}{.402} \\
    \textsc{HUB}$^l$ &  2021 & \cellcolor{w47}{.667} & \cellcolor{w56}{.418} \\
    \textsc{Swin}$^s$ &  2021 & \cellcolor{w50}{.672} &  \cellcolor{w7}{.306} \\
    \textsc{W2V2}$^r$ &  2021 & \cellcolor{w56}{.684} & \cellcolor{w52}{.411} \\
    \textsc{Whisper}$^s$ &  2023 & \cellcolor{w41}{.656} &  \cellcolor{w0}{.279} \\
    \textsc{W2V2}$^e$ &  2023 & \cellcolor{w55}{.684} & \cellcolor{w53}{.411} \\
    \textsc{Whisper}$^t$ &  2023 & \cellcolor{w56}{.684} & \cellcolor{w39}{.380} \\
    \textsc{Whisper}$^b$ &  2023 & \cellcolor{w57}{.686} &  \cellcolor{w0}{.200} \\
  \midrule
  Late Fusion (All)   & - & .676 & .346\\
  Late Fusion (Top-5) & - & .708 & .434\\
\bottomrule
\end{tabular}
\end{table}

\begin{table}[!t]
    \centering
    \caption{
    UAR results for best-performing architectures in the \emph{tuning phase}.
    Showing best results obtained after tuning hyperparameters (batch size, optimiser, learning rate) and keeping the best-performing combination on the official test set.
    Also including $95\%$ confidence intervals for our models computed with bootstrapping.
    SOTA results taken from original works.
    }
    \label{tab:tuning}
\begin{tabular}{lcc}
\toprule
       Model &   $2$-class &   $5$-class \\
\midrule
    2009 Baseline & .677 & .382 \\
    2009 Winners  & .703 & .417 \\
    2009 Fusion   & .712 & .440 \\
    \citet{Zhao19-EDS} & N/A & .454 \\
    \midrule
   \textsc{IS10}$^d$ & .685 [.674 - .696] &   .394 [.377 - .411] \\
   \textsc{Resnet50} & .690 [.680 - .701] &   .423 [.405 - .441] \\
   \textsc{CNN14} & .672 [.661 - .683] &   .448 [.428 - .467] \\
   \textsc{W2V2}$^e$ & \textbf{.717 [.706 - .728]} &   .448 [.431 - .465] \\
   \textsc{Whisper}$^t$ & .707 [.696 - .718] & \textbf{.454 [.437 - .472]} \\
\bottomrule
\end{tabular}
\end{table}

Results for the \emph{exploration phase} are presented in \Cref{tab:exploration}.
The best-performing model for the 2-class problem is \textsc{CNN14}, with an \ac{UAR} of $.692$, whereas for the 5-class problem it is \textsc{ResNet50}, with a \ac{UAR} of $.428$.
We further observe that some models have failed to converge and yield chance (or near-chance) performance -- most likely caused by the choice of hyperparameters, which favour some models more than others.
Notably, most of these results are below the challenge winners (\ac{UAR}: $.703/.417$) and several are in the same `ballpark' as the original baseline (\ac{UAR}: $.677/.382$).
As in \citet{Schuller11-RRE}, a fusion of our top models (here we only take 5) improves performance.

Turning to the \emph{tuning phase}, \cref{tab:tuning} shows the \emph{best} performance for the top five models
%BS: why top five? This needs explanation, please. In the challenge, we did EVERY NUMBER 1, 2, 3... and finally reported the fusion for the N best with N optimised - here, it sounds like a random choice?
from \Cref{tab:exploration} after optimising standard hyperparameters (optimiser, learning rate, batch size).
In this case, \textsc{W2V2}$^e$ yields the best performance for the 2-class problem ($.717$) and \textsc{Whisper}$^t$ for the 5-class problem ($.454$) -- in both cases, we reach results better than the challenge winners, albeit with the gains for the 2-class problem being marginal.
%BS: significance testing? :)
Given that \textsc{Whisper} has been trained for multilingual \ac{ASR} (including German), and that previous performance improvements on valence prediction for English speech heavily depended on implicit linguistic knowledge~\citep{Triantafyllopoulos22-PSE}, we expect \textsc{Whisper}'s success to be also attributed to that aspect. %BS: WHY? FAU AIBO is GERMAN? Whisper is English? Please clarify.

However, it is still the case that all models we have tested remain close to or even below the original challenge baseline and winners, and especially the fusion of the top challenge submissions.
This remains so even after selecting only the best-performing model out of all tested hyperparameters, essentially following a generally bad practice of overfitting.
This was done intentionally to gauge performance under the most optimistic of settings -- that of virtually unrestricted evaluation runs.
We note that the original challenge participants were given $25$ runs each.
This shows
%BS: Why? How is this linked? Because you had less than 25? Please explain.
how the gains we obtain here 
%over the older state-of-the-art 
must be further tempered to account for more runs on our side.
% recent \ac{DNN} advances have produced marginal gains on the FAU-AIBO dataset.

% \ab{we have to point out explicitly that IS09 had no dev set so the winners from IS09 and our best results are comparable but not in a strict sense.)}

\begin{figure}
    \centering
    \includegraphics[width=.45\columnwidth]{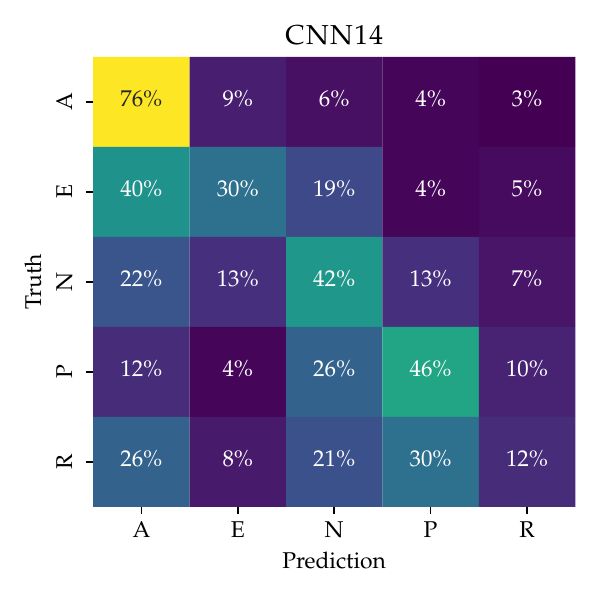}~%
    \includegraphics[width=.45\columnwidth]{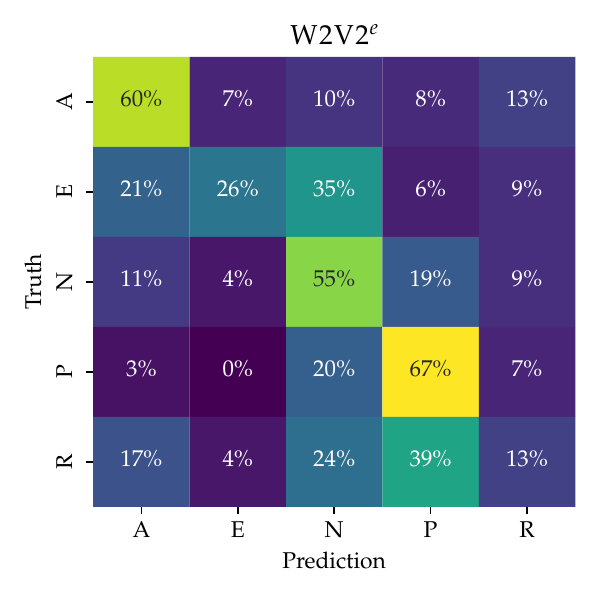}
    \caption{
    Test set confusion matrices (in \%) for the best-performing \textsc{CNN14} and \textsc{W2V2}$^e$ models on the $5$-class problem. %BS: on WHAT PARTITION? Test? Thanks for adding :) Also, we usually do NOT do % in confusion matrices, as it is inexact. The heatmap (missing here) usually serves best for that, but it's ok :)
    }
    \label{fig:confusion}
\end{figure}

\noindent
\textbf{Do newer/larger models perform better?} Interestingly, when looking at the results of the exploration phase, there is no correlation of \ac{UAR} performance with the year of publication (Spearman's $\rho = .12/.09$ for the $2-$/$5-$class problem), and the correlation with the amount of \acp{MAC} and trainable parameters is also very low ($\rho = .15/.23$ and $\rho = -.08/.09$) -- note that model \acp{MAC} and parameters do not account for feature extraction.
This further illustrates how neither more recent nor more complex models are able to surpass the prior state-of-the-art.
Finally, the ranking of model performance between the $2$- and $5$-class problems is moderate ($\rho = .47$); this shows that models are not consistently good when given the same data but different labels (\ie,  our findings are consistent with the standard ``no free-lunch'' theorem).
Surprisingly, some models even show near chance-level performance on one task while performing well on the other.

\noindent
\textbf{Agreement between different models:}
Different models agree with one another to a moderate or good extent.
The average pairwise agreement (percentage of instances where two models agree) for all models of the exploration phase is $70\%$ and $55\%$ for the $2$- and $5$-class models, which rises to $80\%$ and $57\%$, respectively, when considering only the top-5 ones.
Additionally, this is exemplified by considering the confusion matrices 
in \cref{fig:confusion}
of the best-performing \textsc{CNN14} and \textsc{W2V2}$^e$ from the tuning phase -- even though they result in an almost identical \ac{UAR}, their behaviour on the test set is not very similar.
For example, \textsc{CNN14} shows a higher recall for the angry and emphatic classes, to the detriment of more neutral samples misclassified as such.
Overall, this demonstrates that models trained on similar data do not converge to an identical solution -- a finding congruent with the literature on underspecification~\citep{Damour22-UPC}.

%SR: in the following section it may make sense to point out that the data is ranked according to the difficulty score, or at least this is done for CE Loss.
\noindent
\textbf{Model vs human performance:}
We also investigate whether samples that are harder to classify for humans are also harder for models.
The standard FAU-AIBO release comes with annotator confidences per instance, computed by taking the percentage of annotators who agree with the gold standard;
we thus define `difficulty` as $1$ minus that confidence.
We then make the following observations when considering all models of the tuning phase:
% SR: if difficulty is based on the "Predefined" scoring function, then it is ranked descending (i.e. higher == easier)  and then normalized in the range of [0,1]. It should not make a difference when using spearman but i just wanted to note it.

a) We first adopt a model-agnostic measure of difficulty, which we define as the number of models who disagree with the max-vote computed by all models on each instance -- this is akin to the computation of difficulty for the human annotators.
Spearman's $\rho$ between this measure and annotator disagreement is moderate ($.33$ and $.20$ for the $2$- and $5$-class problems).

b) We then adopt a model-specific measure of difficulty, defined as the cross-entropy loss for each instance, similar to \citet{Hacohen19-OTP}.
Different models have different rankings of instance difficulty, with average pairwise $\rho$ being $.51$ and $.33$ for the $2-$ and $5-$class problems.

c) Finally, we compute the Spearman $\rho$ between each model's \ac{UAR} and its Spearman $\rho$ with annotator disagreement;
here, $\rho$ is $-.38/-.07$ for the $2$/$5$-class problem, indicating that larger agreement with human annotators does not lead to better performance (rather, the opposite).
Collectively, our results indicate that models appear to learn differently than humans, with small agreement to what constitutes an easy or hard example.
% On the other hand, the average annotator agreement is $/\%$ for the $2$-/$5$-class problems, thus showing that our
% The `larger picture' implication is that there is still room for improvement, 
% given that we have not yet reached human-level performance. 
%BS: Why? Do we KNOW human-level performance? I think we only would if we had OTHER humans try it on the FULL test data - ideally also given the training material first. Perhaps better rephrase?

\noindent
\textbf{Limitations:}
Our study is obviously limited with respect to the approaches we tried; with hundreds of papers published on \ac{SER} on a yearly basis, it was impossible to evaluate all of them.
We thus opted for the simplest ones: fine-tuning large \acp{DNN} previously shown to be successful on other datasets using a range of standard hyperparameters.
% We have also refrained from explicitly using linguistics and measuring the advances obtained by progress in large language models, something we leave for future work.
Furthermore, we have focused exclusively on one dataset; we intend to explore whether these findings generalise to other datasets in a follow-up work. %BS: Would be great to do the 15-years ComParE article this year :)

% \ab{wenn Platz ist, könnte man doch die Korr.Matritzen von IS09 und dem besten approach  W2v2-l-emo miteinander vergleichen, ob sich die überhaupt wesentlich unterscheiden?
% ~~\\
% das Wort `Aibo' als Vokativ kommt sehr oft vor -- man koennte checken, ob es besser/schlechter modelliert wird -- ist ja linguistisch `neutral'}

\section{Conclusion}
We have conducted a large-scale study of several modern \ac{DNN} architectures -- most of them pre-trained on large datasets -- on the data of the  INTERSPEECH 2009 Emotion Challenge. %BS: this is the only valid title - please do not write differently 
Given standard parameters, we were only able to marginally  outperform the state-of-the-art achieved by challenge participants, with several models scoring even below the baseline, and some failing to converge altogether.
Further optimising hyperparameters led us to outperform the challenge winners by small margins.
Our subsequent analysis showed that performance improvements have not been consistent over time and are not caused by increased model size.
Moreover, we have found that different models converge to different (sometimes complementary) solutions while differing in how challenging they find individual instances compared to human annotators.
Collectively, our findings suggest that recent success achieved by \ac{DNN} models must be tempered, at least when considering the FAU-AIBO setup.
% On a positive note, they also show that there is still room for improvement in order to approach human-level performance.

%\newpage
\section{Acknowledgements}
%BS: took this out - should not be on page 5? Now, we don't need it, anyhow, so let's not. Yes, Stefan acknowledgement could be good for camera.
%BS: Please fill p.5 until end of page w/ references.
%AB: soweit ich weiss, kann acknowl. nicht auf die 5. Seite -- oder weißt du was anderes? Hm, wir könnten auch noch dem Stefan danken bzw. an ihn erinnern, wenn wir nicht mehr anonymisiert sind. @Björn, was meinst du?
%AT: Doch, bei der Vorlage steht "The 5th page is reserved exclusively for acknowledgements and references, which may begin on an earlier page if there is space.", für Stefan: von mir aus ja, gerne, bitte jetzt schon was schreiben damit ich weiß wie viel Platz wir noch haben

%\ifinterspeechfinal
%     The INTERSPEECH 2023 organisers
%\else
%     The authors
%\fi
%Acknowledgements are removed for anonymous submission.
%AB: to be included if accepted: 
This work was funded from the DFG’s Reinhart Koselleck project No.\ 442218748 (AUDI0NOMOUS) and the EU H2020 project No.\ 101135556 (INDUX-R), the Munich Data Science Institute (MDSI), and the Munich Center for Machine Learning (MCML). We would also like to remember our colleague Stefan Steidl -- dedicated co-organiser of the ComParE challenges --  who unexpectedly passed away at young age in October 2018.
%\textcolor{red}{
%This work was funded from the DFG’s Reinhart Koselleck
%project No.\ 442218748 (AUDI0NOMOUS).
%AB: imho, we could/should add: 
%
% ~~\\
% ~~\\
% ~~\\
% ~~\\
% ~~\\
% ~~\\

%}
%would like to thank ISCA and the organising committees of past INTERSPEECH conferences for their help and for kindly providing the previous version of this template.

%As a final reminder, the 5th page is reserved exclusively for references. No other content must appear on the 5th page. Appendices, if any, must be within the first 4 pages. The references may start on an earlier page, if there is space.

% \newpage

% \bibliographystyle{IEEEtran}
% \bibliography{mybib}
\section{\refname}
 \printbibliography[heading=none]

\end{document}